\documentclass[11pt,letterpaper]{article}
\usepackage{emnlp2017}
\usepackage{times}
\usepackage{latexsym}
\usepackage{amssymb}
\usepackage{paralist}
\usepackage{graphicx}
\usepackage{color}
\usepackage{arydshln}
\usepackage{multirow}

\emnlpfinalcopy

\title{Shortcut-Stacked Sentence Encoders for Multi-Domain Inference}

\author{Yixin Nie \and Mohit Bansal \\
        UNC Chapel Hill\\
        \texttt{\{yixin1, mbansal\}@cs.unc.edu}
        }

\date{}

\begin{document}

\maketitle

\begin{abstract}
We present a simple sequential sentence encoder for multi-domain natural language inference. Our encoder is based on stacked bidirectional LSTM-RNNs with shortcut connections and fine-tuning of word embeddings. The overall supervised  model uses the above encoder to encode two input sentences into two vectors, and then uses a classifier over the vector combination to label the relationship between these two sentences as that of entailment, contradiction, or neural. Our Shortcut-Stacked sentence encoders achieve strong improvements over existing encoders on matched and mismatched multi-domain natural language inference (top non-ensemble single-model result in the EMNLP RepEval 2017 Shared Task~\cite{nangia2017repeval}). Moreover, they achieve the new state-of-the-art encoding result on the original SNLI dataset~\cite{snli:emnlp2015}.

\end{abstract}

\section{Introduction and Background}

\begin{figure*}[ht!]
\centering
\includegraphics[width=140mm]{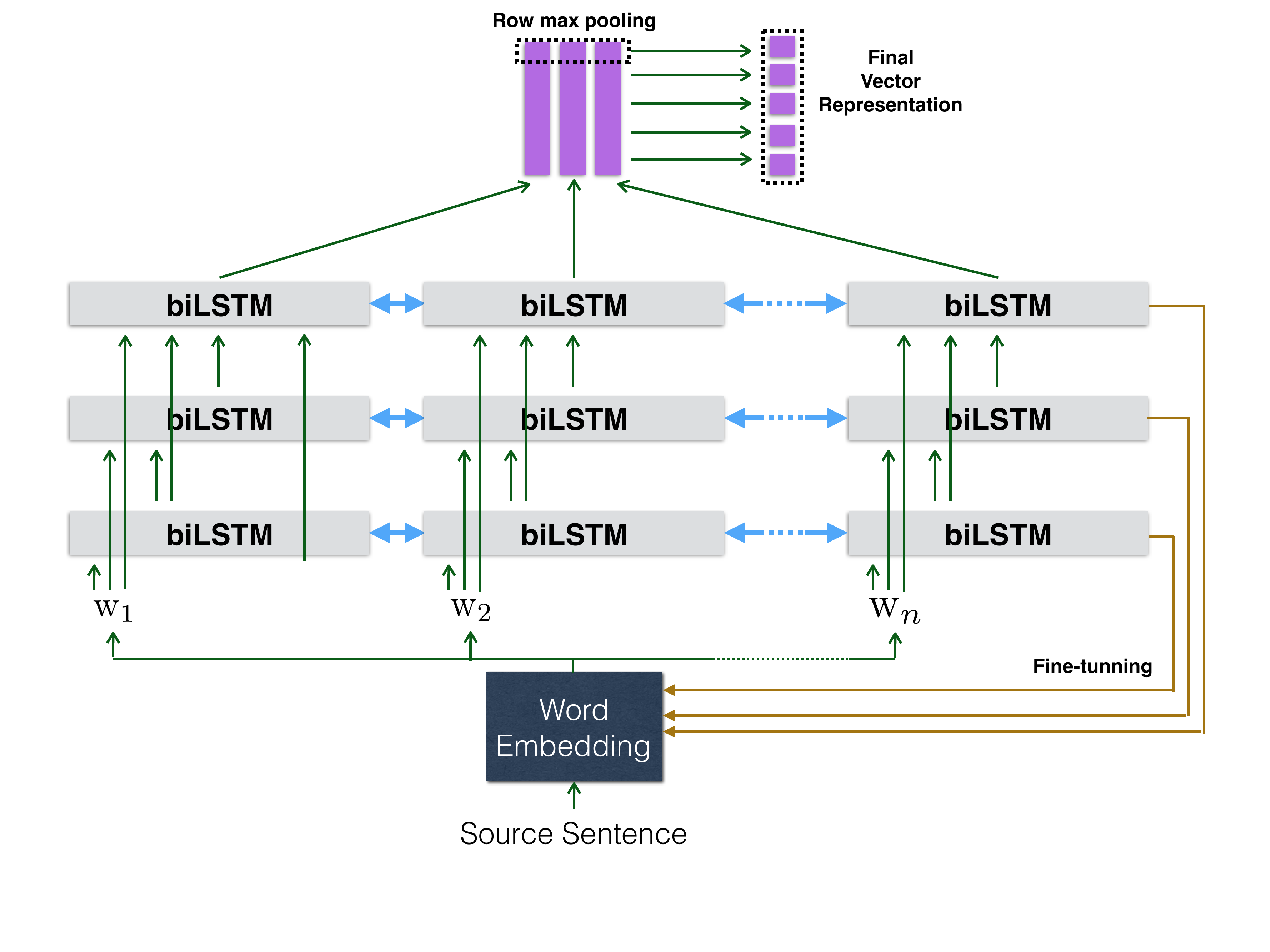}
\vspace{-35pt}
\caption{Our encoder's architecture: stacked biLSTM with shortcut connections and fine-tuning.\label{fig:model}}
\end{figure*}

Natural language inference (NLI) or recognizing textual entailment (RTE) is a fundamental semantic task in the field of natural language processing. The problem is to determine whether a given hypothesis sentence can be logically inferred from a given premise sentence.
Recently released datasets such as the Stanford Natural Language Inference Corpus \cite{snli:emnlp2015} (SNLI) and the Multi-Genre Natural Language Inference Corpus \cite{williams2017broad} (Multi-NLI) have not only encouraged several end-to-end neural network approaches to NLI, but have also served as an evaluation resource for general representation learning of natural language.

Depending on whether a model will first encode a sentence into a fixed-length vector without any incorporating information from the other sentence, 
the several proposed models can be categorized into two groups: (1) encoding-based models (or sentence encoders), such as Tree-based CNN encoders (TBCNN) in~\newcite{mou2015natural} or Stack-augmented Parser-Interpreter Neural Network (SPINN) in~\newcite{bowman2016fast}, and (2) joint, pairwise models that use cross-features between the two sentences to encode them, such as the Enhanced Sequential Inference Model (ESIM) in~\newcite{chen2017enhanced} or the bilateral multi-perspective matching (BiMPM) model~\newcite{wang2017bilateral}. Moreover, common sentence encoders can again be classified into tree-based encoders such as SPINN in~\newcite{bowman2016fast} which we mentioned before, or sequential encoders such as the biLSTM model by~\newcite{snli:emnlp2015}.

In this paper, we follow the former approach of encoding-based models, and propose a novel yet simple sequential sentence encoder for the Multi-NLI problem. Our encoder does not require any syntactic information of the sentence. It also does not contain any attention or memory structure. It is basically a stacked (multi-layered) bidirectional LSTM-RNN with shortcut connections (feeding all previous layers' outputs and word embeddings to each layer) and word embedding fine-tuning. The overall supervised  model uses these shortcut-stacked encoders to encode two input sentences into two vectors, and then we use a classifier over the vector combination to label the relationship between these two sentences as that of entailment, contradiction, or neural (similar to the classifier setup of~\newcite{snli:emnlp2015} and ~\newcite{conneau2017supervised}). 
Our simple shortcut-stacked encoders achieve strong improvements over existing encoders due to its multi-layered and shortcut-connected properties, on both matched and mismatched evaluation settings for multi-domain natural language inference, as well as on the original SNLI dataset. It is the top single-model (non-ensemble) result in the EMNLP RepEval 2017 Multi-NLI Shared Task~\cite{nangia2017repeval}, and the new state-of-the-art for encoding-based results on the SNLI dataset~\cite{snli:emnlp2015}.

\noindent{\textbf{Github Code Link:} \url{https://github.com/easonnie/multiNLI_encoder}}

\section{Model}
Our model mainly consists of two separate components, a sentence encoder and an entailment classifier. The sentence encoder compresses each source sentence into a vector representation and the classifier makes a three-way classification based on the two vectors of the two source sentences. The model follows the `encoding-based rule', i.e., the encoder will encode each source sentence into a fixed length vector without any information or function based on the other sentence (e.g., cross-attention or memory comparing the two sentences). In order to fully explore the generalization of the sentence encoder, the same encoder is applied to both the premise and the hypothesis with shared parameters projecting them into the same space. This setting follows the idea of Siamese Networks in~\newcite{bromley1994signature}. Figure~\ref{fig:model} shows the overview of our encoding model (the standard classifier setup is not shown here; see~\newcite{snli:emnlp2015} and ~\newcite{conneau2017supervised} for that).

\subsection{Sentence Encoder}
Our sentence encoder is simply composed of multiple stacked bidirectional LSTM (biLSTM) layers with shortcut connections followed by a max pooling layer.
Let bilstm$^i$ represent the $i$th biLSTM layer, which is defined as:
\begin{equation}
h_t^i = \mathrm{bilstm}^i(x_t^i,t),\forall t \in [1,2,...,n]
\end{equation}
where $h_t^i$ is the output of the $i$th biLSTM at time t over input sequence $(x_1^i, x_2^i, ..., x_n^i)$.

In a typical \textbf{stacked biLSTM} structure, the input of the next LSTM-RNN layer is simply the output sequence of the previous LSTM-RNN layer. In our settings, the input sequences for the $i$th biLSTM layer are the concatenated outputs of \emph{all the previous layers}, plus the original word embedding sequence. This gives a \textbf{shortcut connection} style setup, related to the widely used idea of residual connections in CNNs for computer vision~\cite{he2016deep}, highway networks for RNNs in speech processing~\cite{zhang2016highway}, and shortcut connections in hierarchical multitasking learning~\cite{hashimoto2016joint}; but in our case we feed in all the previous layers' output sequences as well as the word embedding sequence to every layer.

Let $W=(w_1, w_2, ..., w_n)$ represent words in the source sentence. We assume $w_i \in \mathbb{R}^d$ is a word embedding vector which are initialized using some pre-trained vector embeddings (and is then fine-tuned end-to-end via the NLI supervision). Then, the input of $i$th biLSTM layer at time $t$ is defined as:
\begin{equation}
x_t^1 = w_t
\end{equation}
\begin{equation}
x_t^i = [w_t, h_t^{i-1}, h_t^{i-2}, ... h_t^{1}] \quad (\textrm{for} \;\; i>1)\\
\end{equation}
where $[]$ represents vector concatenation.

Then, assuming we have $m$ layers of biLSTM, the final vector representation will be obtained by applying row-max-pool over the output of the last biLSTM layer, similar to~\newcite{conneau2017supervised}. The final layer is defined as:
\begin{equation}
H^m=(h_1^m, h_2^m, ..., h_n^m)
\end{equation}
\begin{equation}
v=max(H^m)
\end{equation}
where $h_i^m,v \in \mathbb{R}^{2d_m}$, $H^m \in \mathbb{R}^{2d_m \times n}$, $d_m$ is the dimension of the hidden state of the last forward and backward LSTM layers, and $v$ is the final vector representation for the source sentence (which is later fed to the NLI classifier).

The closest encoder architecture to ours is that of~\newcite{conneau2017supervised}, whose model consists of a single-layer biLSTM with a max-pooling layer, which we treat as our starting point. Our experiments (Section~\ref{sec:result}) demonstrate that  our enhancements of the stacked-biRNN with shortcut connections provide significant gains on top of this baseline (for both SNLI and Multi-NLI).

\subsection{Entailment Classifier}
After we obtain the vector representation for the premise and hypothesis sentence, we apply three matching methods to the two vectors \begin{inparaenum}[(i)]
\item{concatenation}
\item{element-wise distance} and
\item{element-wise product}
\end{inparaenum} for these two vectors and then concatenate these three match vectors (based on the heuristic matching presented in~\newcite{mou2015natural}). Let $v_p$ and $v_h$ be the vector representations for premise and hypothesis, respectively. The matching vector is then defined as:
\begin{equation}
m=[v_p, v_h, \left|v_p - v_h\right|, v_p \otimes v_h]
\end{equation}
At last, we feed this final concatenated result $m$ into a MLP layer and use a softmax layer to make final classification.

\begin{table}[ht!]
\begin{center}
\small
\begin{tabular}{|c|c|cc|}
\hline
\multicolumn{2}{|c|}{Layers and Dimensions} & \multicolumn{2}{|c|}{Accuracy} \\
\hline
\#layers & bilstm-dim & Matched & Mismatched \\
\hline
1 & 512 & 72.5 & 72.9\\
2 & 512 + 512 & 73.4 & 73.6\\
1 & 1024 & 72.9 & 72.9\\
2 & 512 + 1024 & 73.7 & 74.2\\
1 & 2048 & 73.0 & 73.5\\
2 & 512 + 2048 & 73.7 & 74.2\\
2 & 1024 + 2048 & 73.8 & 74.4\\
2 & 2048 + 2048 & 74.0 & 74.6\\
3 & 512 + 1024 + 2048 & \textbf{74.2} & \textbf{74.7}\\
\hline
\end{tabular}
\end{center}
\caption{Analysis of results for models with different \# of biLSTM layers and their hidden state dimensions.
}\label{tab:lstmlayersdims}
\end{table}

\begin{table}[ht!]
\begin{center}
\small
\begin{tabular}{|c|cc|}
\hline
& Matched & Mismatched \\
\hline
without any shortcut connection & 72.6 & 73.4\\
only word shortcut connection & 74.2 & 74.6\\
full shortcut connection & \textbf{74.2} & \textbf{74.7}\\
\hline
\end{tabular}
\end{center}
\caption{Ablation results with and without shortcut connections.}\label{tab:residual}
\end{table}

\begin{table}[ht!]
\begin{center}
\small
\begin{tabular}{|c|cc|}
\hline
Word-Embedding & Matched & Mismatched \\
\hline
fixed & 71.8 & 72.6\\
fine-tuned & \textbf{72.7} & \textbf{72.8}\\
\hline
\end{tabular}
\end{center}
\caption{Ablation results with and without fine-tuning of word embeddings.}\label{tab:ftwordembedding}
\end{table}

\begin{table}[ht!]
\begin{center}
\small
\begin{tabular}{|cc|cc|}
\hline
\# of MLPs & Activation & Matched & Mismatched \\
\hline
1 & tanh & 73.7 & 74.1\\
2 & tanh & 73.5 & 73.6\\
1 & relu & 74.1 & 74.7\\
2 & relu & \textbf{74.2} & \textbf{74.7}\\
\hline
\end{tabular}
\end{center}
\caption{Ablation results for different MLP classifiers.}\label{tab:classifier}
\end{table}

\section{Experimental Setup}
\subsection{Datasets}
As instructed in the RepEval Multi-NLI shared task, we use all of the training data in Multi-NLI combined with 15\% randomly selected samples from the SNLI training set resampled at each epoch) as our final training set for all models; and we use both the cross-domain (`mismatched') and in-domain (`matched') Multi-NLI development sets for model selection. For the SNLI test results in Table~\ref{tab:encoder_snli}, we train on only the SNLI training set (and we also verify that the tuning decisions hold true on the SNLI dev set).

\begin{table*}[ht!]
\begin{center}
\small
\begin{tabular}{|c|ccc|}
\hline
\multirow{2}{*}{\bf Model} & \multicolumn{3}{|c|}{\bf Accuracy} \\
\ & {\bf SNLI} & {\bf Multi-NLI Matched} & {\bf Multi-NLI Mismatched} \\
\hline
CBOW~\cite{williams2017broad} & 80.6 & 65.2 & 64.6 \\
biLSTM Encoder~\cite{williams2017broad} & 81.5 & 67.5 & 67.1 \\
300D Tree-CNN Encoder~\cite{mou2015natural} & 82.1 & -- & -- \\
300D SPINN-PI Encoder~\cite{bowman2016fast} & 83.2 & -- & -- \\
300D NSE Encoder~\cite{munkhdalai2016neural} & 84.6 & -- & -- \\
biLSTM-Max Encoder~\cite{conneau2017supervised} & 84.5 & -- & -- \\
\hline
Our biLSTM-Max Encoder & 85.2 & 71.7 & 71.2 \\
Our Shortcut-Stacked Encoder & {\bf 86.1} & {\bf 74.6} & {\bf 73.6}\\
\hline
\end{tabular}
\end{center}
\vspace{-7pt}
\caption{Final Test Results on SNLI and Multi-NLI datasets.
}
\vspace{-2pt}
\label{tab:encoder_snli}
\end{table*}

\subsection{Parameter Settings}
We use cross-entropy loss as the training objective with Adam-based ~\cite{kingma2014adam} optimization with 32 batch size. The starting learning rate is 0.0002 with half decay every two epochs. The number of hidden units for MLP in classifier is 1600. Dropout layer is also applied on the output of each layer of MLP, with dropout rate set to 0.1. We used pre-trained 300D Glove 840B vectors~\cite{pennington2014glove} to initialize the word embeddings. Tuning decisions for word embedding training strategy, the hyperparameters of dimension and number of layers for biLSTM, and the activation type and number of layers for MLP, are all explained in Section \ref{sec:result}.

\section{Results and Analysis}
\label{sec:result}

\subsection{Ablation Analysis Results}
We now investigate the effectiveness of each of the enhancement components in our overall model. These ablation results are shown in Tables~\ref{tab:lstmlayersdims},~\ref{tab:residual},~\ref{tab:ftwordembedding} and~\ref{tab:classifier}, all based on the Multi-NLI development sets. Finally, Table~\ref{tab:encoder_snli} shows results for different encoders on SNLI and Multi-NLI test sets.

First, Table~\ref{tab:lstmlayersdims} shows the performance changes for different number of biLSTM layers and their varying dimension size. The dimension size of a biLSTM layer is referring to the dimension of the hidden state for both the forward and backward LSTM-RNNs. As shown, each added layer model improves the accuracy and we achieve a substantial improvement in accuracy (around 2\%) on both matched and mismatched settings, compared to the single-layer biLSTM in~\newcite{conneau2017supervised}. We only experimented with up to 3 layers with 512, 1024, 2048 dimensions each, so the model still has potential to improve the result further with a larger dimension and more layers.

Next, in Table~\ref{tab:residual}, we show that the shortcut connections among the biLSTM layers is also an important contributor to accuracy improvement (around 1.5\% on top of the full 3-layered stacked-RNN model). This demonstrates that simply stacking the biLSTM layers is not sufficient to handle a complex task like Multi-NLI and it is significantly better to have the higher layer connected to both the output and the original input of all the previous layers (note that Table~\ref{tab:lstmlayersdims} results are based on multi-layered models with shortcut connections).

Next, in Table~\ref{tab:ftwordembedding}, we show that fine-tuning the word embeddings also improves results, again for both the in-domain task and cross-domain tasks (the ablation results are based on a smaller model with a 128+256 2-layer biLSTM). Hence, all our models were trained with word embeddings being fine-tuned. 
The last ablation in Table~\ref{tab:classifier} shows that a classifier with two layers of relu is preferable than other options. Thus, we use that setting for our strongest encoder.

\subsection{Multi-NLI and SNLI Test Results}
Finally, in Table~\ref{tab:encoder_snli}, we report the test results for MNLI and SNLI. 
First for Multi-NLI, we improve substantially over the CBOW and biLSTM Encoder baselines reported in the dataset paper~\cite{williams2017broad}. We also show that our final shortcut-based stacked encoder achieves around 3\% improvement as compared to the 1-layer biLSTM-Max Encoder in the second last row (using the exact same classifier and optimizer settings). Our shortcut-encoder was also the top singe-model (non-ensemble) result on the EMNLP RepEval Shared Task leaderboard.

Next, for SNLI, we compare our shortcut-stacked encoder with the current state-of-the-art  encoders from the SNLI leaderboard (\url{https://nlp.stanford.edu/projects/snli/}). We also compare to the recent biLSTM-Max Encoder of~\newcite{conneau2017supervised}, which served as our model's 1-layer starting point.\footnote{Note that the `Our biLSTM-Max Encoder' results in the second-last row are obtained using our reimplementation of the ~\newcite{conneau2017supervised} model; our version is 0.7\% better, likely due to our classifier and optimizer settings.}  
The results indicate that `Our Shortcut-Stacked  Encoder' surpasses all the previous state-of-the-art encoders, and achieves the new best encoding-based result on SNLI, suggesting the general effectiveness of simple shortcut-connected stacked layers in sentence encoders.

\section{Conclusion}
\label{sec:conclusion_and_future_work}
We explored various simple combinations and connections of biLSTM-RNN layered architectures and developed a Shortcut-Stacked Sentence Encoder for natural language inference. Our model is the top single result in the EMNLP RepEval 2017 Multi-NLI Shared Task, and it also surpasses the state-of-the-art encoders for the SNLI dataset. In future work, we are also evaluating the effectiveness of shortcut-stacked sentence encoders on several other semantic tasks.

\section{Addendum: Shortcut vs. Residual}
In later experiments, we found that a residual connection can achieve similar accuracies with fewer number of parameters, compared to a shortcut connection. Therefore, in order to reduce the model size and to also follow the SNLI leader-board settings (e.g., 300D and 600D embeddings), we performed some additional SNLI experiments with the shortcut connections replaced with residual connections, where the input to each next biLSTM layer is the concatenation of the word embedding and the summation of outputs of all previous layers (related to ResNet in computer vision~\cite{he2016deep}).
Table~\ref{tab:test_snli_table} shows these residual-connection SNLI test results and the parameter comparison to shortcut-connection models (using 3 stacked-biLSTM layers, and one 800-unit MLP layer, based on SNLI dev set tuning).

\begin{table}[ht]
\begin{center}
\small
\begin{tabular}{|c|ccc|}
\hline
{\bf Model} & {\bf \#param} & {\bf Dev}  & {\bf Test} \\
\hline
300D Residual-Stacked-Encoder & 9.7M & 86.4 & 85.7\\
600D Residual-Stacked-Encoder & 28.9M & \textbf{87.0} & \textbf{86.0}\\
\hline
600D Shortcut-Stacked-Encoder & 34.7M & 86.8 & 85.9\\
\hline
\end{tabular}
\end{center}
\caption{Results on SNLI for the fewer-parameter Residual-Stacked Encoder models. Each model has 3 biLSTM-stacked layers and 1 MLP layer. The \#param column denotes the number of parameters in millions.}\label{tab:test_snli_table}
\end{table}

\section*{Acknowledgments}
We thank the shared task organizers and the anonymous reviewers. This work was partially supported by a Google
Faculty Research Award, an IBM Faculty Award,
a Bloomberg Data Science Research Grant, and
NVidia GPU awards.


\bibliography{emnlp2017}
\bibliographystyle{emnlp_natbib}

\end{document}